%% file: iclr2027_conference.tex
\documentclass{article} 
\usepackage{iclr2027_conference,times}

\input{math_commands.tex}

\usepackage{hyperref}
\usepackage{url}

\usepackage{algorithm}
\usepackage{algpseudocode}
\usepackage{booktabs}
\usepackage{capt-of}
\usepackage{graphicx}
\usepackage{multirow}
\usepackage{wrapfig}
\usepackage[table]{xcolor}
\usepackage{enumitem}

\usepackage[most]{tcolorbox}
\usepackage{listings}
\usepackage{xcolor}

\lstdefinestyle{promptstyle}{
  basicstyle=\ttfamily\small,
  columns=fixed,
  basewidth=0.52em,
  breaklines=true,
  breakindent=0pt,
  breakautoindent=false,
  keepspaces=true,
  showstringspaces=false
}

\newtcblisting{promptbox}[1]{
  enhanced,
  listing only,
  listing engine=listings,
  listing options={style=promptstyle},
  colback=black!2,
  colframe=black!20,
  boxrule=0.5pt,
  arc=1mm,
  title={#1},
  fonttitle=\bfseries\small,
  coltitle=black
}

\title{\methodology: Time Series Forecasting with\\Experience-Informed Covariate Judgments}

\author{%
\textbf{Donguk Kwon}$^{1}$\quad
\textbf{Wooseok Jeong}$^{2}$\quad
\textbf{Dongha Lee}$^{1}$\thanks{Corresponding author.}\\
$^{1}$Yonsei University \quad
$^{2}$Konkuk University\\
\texttt{\{donguk.kwon, donalee\}@yonsei.ac.kr} \quad
\texttt{jws010825@konkuk.ac.kr}
}

\newcommand{\methodology}{JudgeCast}

\iclrfinalcopy 
\begin{document}

\maketitle

\vspace{-1.2em}
\begin{abstract}
\input{latex/000_abstract}
\end{abstract}

\vspace{-0.6em}
\section{Introduction}
\vspace{-0.5em}
\input{latex/100_introduction}

\section{Related Work}
\input{latex/200_related_work}

\section{\methodology}
\label{sec:methodology}
\input{latex/300_methodology}

\section{Experiments}
\label{sec:experiments}
\input{latex/400_experiments}

\section{Conclusion}
\input{latex/500_conclusion}

\bibliography{reference}
\bibliographystyle{iclr2027_conference}

\newpage
\appendix
\input{latex/600_appendix}

\end{document}

%% file: math_commands.tex
\usepackage{amsmath,amsfonts,bm}

\def\eqref#1{equation~\ref{#1}}

\def\1{\bm{1}}

\DeclareMathAlphabet{\mathsfit}{\encodingdefault}{\sfdefault}{m}{sl}
\SetMathAlphabet{\mathsfit}{bold}{\encodingdefault}{\sfdefault}{bx}{n}



%% file: latex/000_abstract.tex
Covariate effects vary across contexts and shift over time, requiring forecasters to assess how to use them for each forecasting context.
As forecasting proceeds, observations for earlier forecasts become available, providing feedback on past covariate use for subsequent forecasts.
However, when multiple covariates act together, the forecast error reveals the numerical discrepancy from the observation but not how the covariates should have been used.
We introduce \methodology, an experience-based framework for time series forecasting with covariates.
Following the judgmental adjustment practice, a frozen TSFM provides the base forecast, while a frozen LLM uses the current context and relevant experience to adjust it.
Within the adjustment, assessing covariate effects and determining the numerical adjustment serve distinct roles, so \methodology{} first forms explicit covariate-wise judgments and then determines the adjustment.
After observation, \methodology{} uses the observed residual of the base forecast to reconstruct alternative judgments and evaluates the original and alternatives through their resulting adjustments.
The best-performing decision is selected and retained as validated experience for subsequent forecasts.
Across diverse real-world datasets, \methodology{} outperforms strong baselines.
Ablations show that explicit covariate-wise judgment can improve forecast-time adjustment, while residual-guided experience construction yields more reliable forecasting gains than retaining raw decisions as experience.

%% file: latex/100_introduction.tex
Time series forecasting supports decision-making across a wide range of domains, including healthcare, demand planning, and transportation~\citep{saleh2025healthcare,feddersen2025demand,zhang2026traffic}.
This task has advanced rapidly with time series foundation models (TSFMs), which are pretrained on large collections of series and produce strong zero-shot forecasts for unseen targets~\citep{das2024timesfm,ansari2024chronos,woo2024moirai,ansari2025chronos2}.
Beyond the temporal patterns in the target's own past, its future may also depend on covariates.
The effects of these covariates vary across contexts and shift over time, requiring forecasters to assess how to use them for each forecasting context~\citep{hastie1993varying,foekens1998varying,huang2019structural}.

Existing forecasting approaches address this by incorporating contextual information during forecast generation.
Covariate-informed forecasters model covariate effects together with the target's temporal patterns in a predictive function~\citep{lim2021tft,wang2024timexer,ansari2025chronos2,yang2026baguan}, while large language model (LLM)-based forecasters embed contextual reasoning within forecast generation~\citep{zhou2026timer1,wang2026vot,das2026nexus}.
As forecasting proceeds, observations corresponding to earlier forecasts become available, allowing those forecasts to be evaluated through their forecast errors.
This evaluation provides feedback that can be incorporated into subsequent forecasts through retraining or parameter-efficient adaptation.
Applying these methods repeatedly, however, incurs additional computational cost and latency.
These methods are also unavailable when model parameters cannot be updated.
This motivates using forecast feedback to inform subsequent forecasts without parameter updates.

\begin{figure}[t]
\centering
\includegraphics[width=\linewidth]{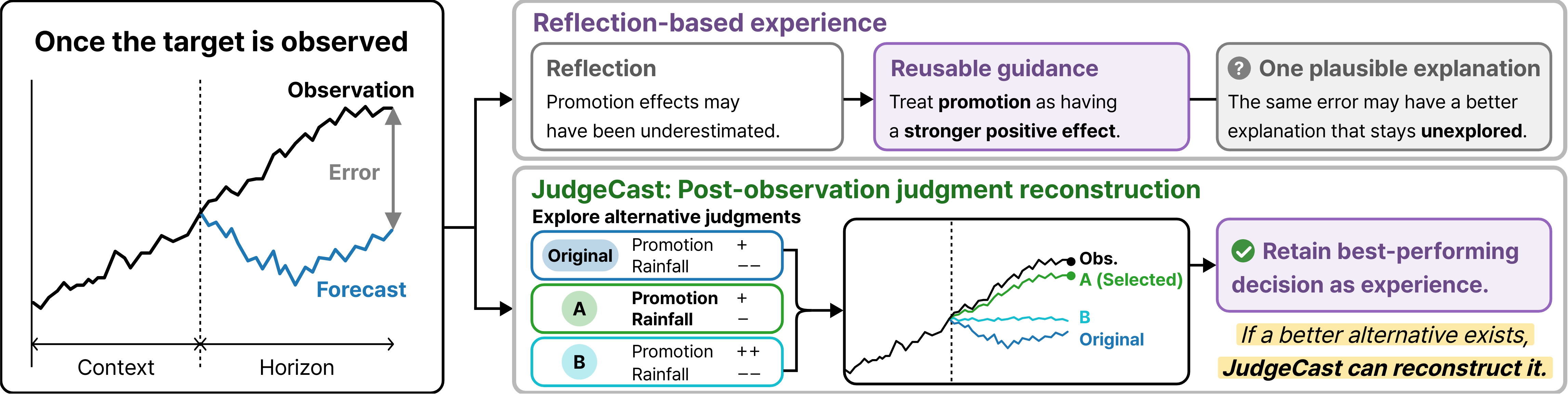}
\vspace{-10pt}
\caption{
Illustration of post-observation judgment reconstruction.
A single reflection provides one plausible explanation, whereas \methodology{} revisits forecast-time judgments by exploring alternatives and comparing their resulting adjustments to construct reusable experience.
}
\vspace{-10pt}
\label{fig:introduction}
\end{figure}

Recent time series forecasting methods use such feedback without parameter updates by deriving reflections or guidance from forecast errors and retaining them as reusable experience for subsequent forecasts~\citep{wang2024newsforecast,tao2026memcast,liao2026lastmile}.
However, when multiple covariates act together, the forecast error does not uniquely indicate how their effects should have been assessed.
For example, if demand exceeds the forecast during a rainy promotion period, the promotion may have been underestimated, the dampening effect of rain may have been overestimated, or rain may have altered how customers responded to the promotion, as illustrated in Figure~\ref{fig:introduction}.
Thus, the forecast error reveals the numerical discrepancy from the observation, but not how the covariates should have been used.
The remaining challenge is therefore to construct reusable experience about covariate use from such forecast feedback.

We introduce \textbf{\methodology{}}, an experience-based framework for time series forecasting with covariates.
At forecast time, \methodology{} formulates covariate-informed forecasting as a judgmental adjustment to a base forecast.
Judgmental adjustment is a forecasting practice in which a practitioner uses contextual information to revise a statistical forecast~\citep{lawrence2006judgmental,fildes2009effective}.
Following this practice, a frozen TSFM produces the base forecast from the target's past alone, while a frozen LLM uses the current context and relevant experience to adjust it.
Within this adjustment, assessing covariate effects from contextual information and determining the numerical adjustment serve distinct roles.
The former assesses the expected effect of each covariate relative to the base forecast, whereas the latter uses these assessments collectively to determine the numerical adjustment.
\methodology{} reflects this distinction by representing the expected covariate effects as explicit covariate-wise judgments and using these judgments to determine the numerical adjustment.
Together, the judgments and adjustment constitute the forecast-time decision.

Once the target observation becomes available, \methodology{} revisits the covariate-wise judgments formed at forecast time to construct experience for subsequent forecasts.
The residual of the base forecast indicates the numerical discrepancy from the observation, but does not determine how the covariate effects should have been judged.
Using this residual as feedback, \methodology{} reconstructs alternative covariate-wise judgments and evaluates the original and alternative judgments through their resulting adjustments.
The best-performing decision is retained as validated experience only if its adjusted forecast improves on the base forecast.
This process uses the observed residual to construct reusable experience about covariate use beyond the forecast error itself.
This validation does not establish that the selected judgment faithfully reflects the effects of the covariates.

Across diverse real-world datasets, \methodology{} outperforms the strongest baselines in our main evaluation by 16.7\% in MSE and 6.5\% in MAE on average.
Ablations show that explicit covariate-wise judgment can improve forecast-time adjustment, while residual-guided experience construction consistently outperforms raw experience by reconstructing past decisions into validated experience.

\begin{itemize}[leftmargin=1.5em, labelsep=0.5em, itemsep=1pt]
\item
We introduce \methodology, an experience-based judgmental adjustment framework for time series forecasting with covariates.
A TSFM provides the base forecast, while an LLM uses relevant experience to form explicit covariate-wise judgments and determine the numerical adjustment.
\item
We develop residual-guided experience construction that reconstructs alternative forecast-time judgments after observation.
The original and alternative judgments are evaluated through their resulting adjustments, and the best-performing decision is retained as validated experience only when its adjusted forecast improves on the base forecast.
\item
We demonstrate consistent forecasting gains across diverse real-world datasets.
Ablations further show that explicit covariate-wise judgment can improve forecast-time adjustment, while residual-guided construction yields more reliable gains than raw experience.
\end{itemize}

%% file: latex/200_related_work.tex
\paragraph{Covariate-Informed Time Series Forecasting.}

Existing covariate-informed forecasters feed covariates into the predictive function together with the target~\citep{lim2021tft,wang2024timexer,ansari2025chronos2,yang2026baguan}, with recent variants guiding this conditioning with statistical priors or external contextual knowledge~\citep{cheng2026kite,wu2026exotimer}.
Covariate effects are thus modeled within forecast generation rather than represented separately as covariate-wise judgments.
Post-hoc attribution quantifies covariate contributions after prediction rather than as a forecast-time decision~\citep{lundberg2017shap,hertel2026explainable}.

\paragraph{LLM-based Time Series Forecasting with Experience.}

Some LLM-based forecasters retrieve past cases to inform later forecasts~\citep{yang2025timerag,wang2026vot}.
Others use forecast errors to refine contextual reasoning or derive forecasting guidance~\citep{wang2024newsforecast,das2026nexus}.
MemCast~\citep{tao2026memcast} further organizes prediction outcomes, inference trajectories, and temporal features into hierarchical forecasting experience for subsequent forecasts.
These approaches reuse past cases, forecast errors, reflections, guidance, or inference trajectories, but do not use the observation to reconstruct the forecast-time decision about how contextual information was used.

\paragraph{Judgmental Adjustment in Time Series Forecasting.}

Automated judgmental adjustment keeps the forecaster fixed and places a separate adjustment over its forecast.
Existing approaches use agents to revise forecasts using external information and reflect after observation~\citep{liao2026lastmile}, train LLMs to revise or refine forecasts~\citep{liu2026posttime,you2026loftllm}, use LLM-guided residual learning to correct frozen backbone forecasts~\citep{kim2026ctrl}, or retrieve similar historical series for post-hoc revision~\citep{liu2025pir}.
Retrieval-based conformal methods similarly retrieve relevant past residuals, but for interval calibration~\citep{heurich2026rarecp,jin2026rccp}.
These approaches automate forecast revision without explicitly separating covariate effect assessment from numerical adjustment.
\methodology{} instead represents expected covariate effects as explicit covariate-wise judgments and uses them to determine the adjustment.

%% file: latex/300_methodology.tex
\begin{figure}[t]
\centering
\includegraphics[width=\linewidth]{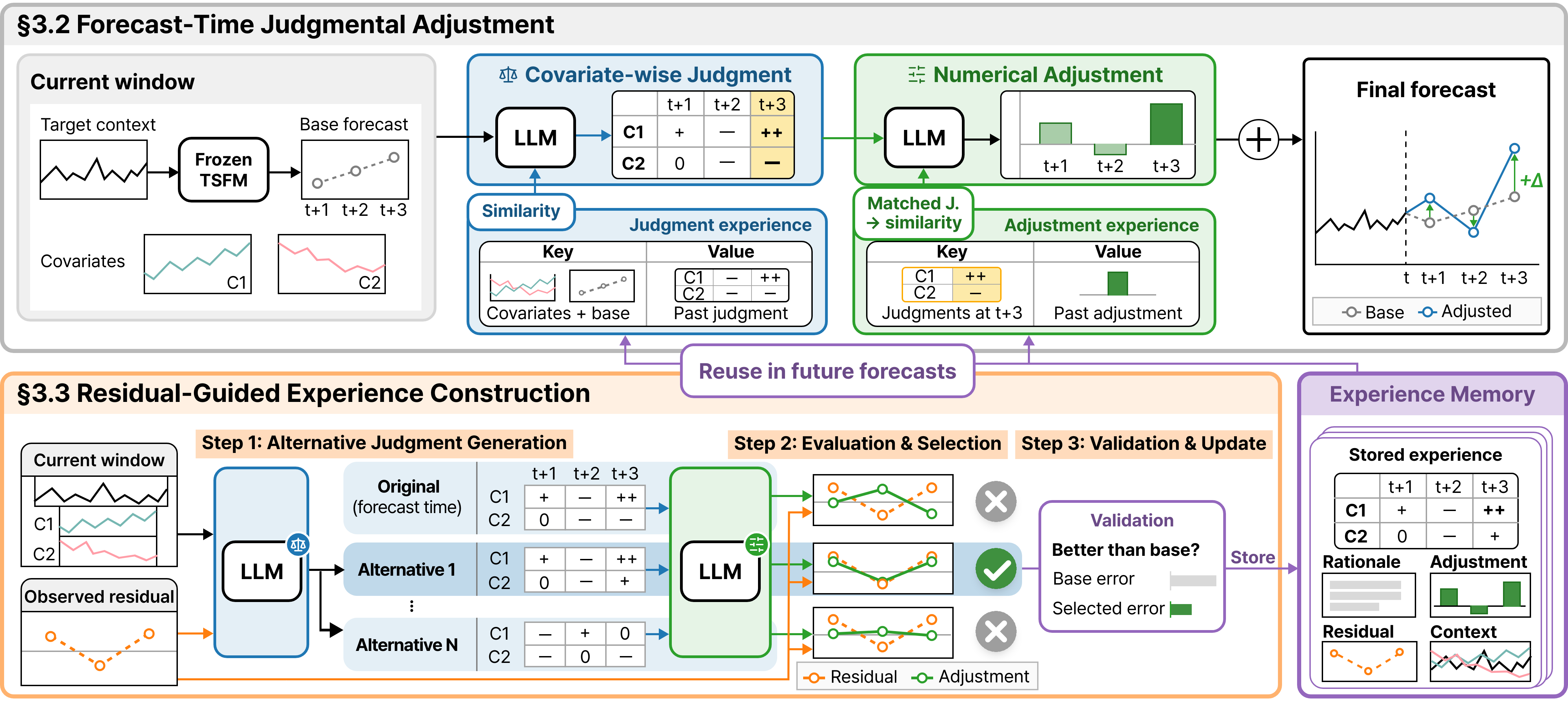}
\vspace{-14pt}
\caption{
Overview of a single forecasting window in \methodology{}.
A frozen TSFM provides the base forecast, while a frozen LLM uses relevant experience to form explicit covariate-wise judgments and determine the numerical adjustment.
After observation, the residual guides alternative judgment generation and adjustment evaluation to select the best-performing decision, which is added to memory as validated experience only when its adjusted forecast improves on the base forecast.
}
\vspace{-7pt}
\label{fig:methodology}
\end{figure}

\methodology{} is an experience-based framework for time series forecasting with covariates over sequential forecasting windows.
It formulates covariate-informed forecasting as a judgmental adjustment to a base forecast.
For each window, a TSFM first produces the base forecast from the target's past alone, while an LLM uses relevant experience to form explicit covariate-wise judgments and determine a numerical adjustment.
Once the target is observed, \methodology{} revisits the forecast-time decision to construct validated experience, which is stored only when the resulting adjusted forecast improves on the base forecast.
Figure~\ref{fig:methodology} illustrates this process for a single forecasting window.

\subsection{Problem Formulation}

Let $y$ denote the target time series and $X=(x_1,\ldots,x_C)$ collect the $C$ covariate time series, where $x_c$ denotes the $c$-th covariate series.
Forecasting proceeds over sequential forecasting windows indexed by $w$, where $y_{1:L}$ is the target context of length $L$, and $y_{L+1:L+H}$ is the target over the forecast horizon of length $H$.
For each window $w$, let $X_w$ denote the covariate values available in that window. We refer to $y_{1:L}$ and $X_w$ collectively as the current context.

Following the practice of judgmental adjustment~\citep{lawrence2006judgmental,fildes2009effective}, we formulate forecasting in each window with a separate base forecast and numerical adjustment.
Let $f_\phi$ and $g_\theta$ denote the frozen TSFM and LLM with parameters $\phi$ and $\theta$, respectively.
The TSFM produces the base forecast from the target context alone,
\begin{equation}
\hat{y}^{\mathrm{base}}
=
f_\phi(y_{1:L}),
\qquad
\hat{y}^{\mathrm{base}}\in\mathbb{R}^{H}.
\end{equation}
For the adjustment, the LLM $g_\theta$ forms explicit covariate-wise judgments and uses them to determine a numerical adjustment $a\in\mathbb{R}^{H}$ to the base forecast.
The final forecast is
\begin{equation}
\hat{y}
=
\hat{y}^{\mathrm{base}} + a,
\qquad
\hat{y}\in\mathbb{R}^{H}.
\end{equation}

\subsection{Forecast-Time Judgmental Adjustment}

At forecast time, assessing covariate effects and determining the numerical adjustment serve distinct roles.
\methodology{} represents each covariate's expected effect relative to the base forecast using pre-defined qualitative judgment labels.
The numerical adjustment is then determined by considering these covariate-wise judgments collectively.
Together, the judgments and adjustment constitute the forecast-time decision.

Let $\mathcal{S}=\{--,-,0,+,++\}$ denote the set of judgment labels.
For covariate $x_c$ and forecast step $h$, $J_{c,h}\in\mathcal{S}$ denotes the direction and qualitative strength of its expected effect relative to the base forecast.
The signs $+$ and $-$ indicate upward and downward effects, repeated signs indicate greater strength, and $0$ denotes no meaningful effect.
Collectively, the judgments form $J\in\mathcal{S}^{C\times H}$.

To ground the current forecast-time decision in preceding windows, \methodology{} uses experience from those windows and their associated decisions.
Let $\mathcal{M}_{<w}$ denote the memory of validated experience accumulated from preceding windows.
To inform the current judgments, \methodology{} retrieves relevant experience based on similarity in the normalized covariate values and base forecast.
Let $\mathcal{E}^{\mathrm{jud}}_w\subseteq\mathcal{M}_{<w}$ denote the retrieved relevant experience.
The LLM forms the current judgments as
\begin{equation}
J
=
g_\theta^{\mathrm{jud}}
\left(
y_{1:L},
X_w,
\hat{y}^{\mathrm{base}},
\mathcal{E}^{\mathrm{jud}}_w
\right).
\end{equation}
Superscripts distinguish the roles of the same LLM $g_\theta$ across the procedure.

For numerical adjustment, \methodology{} determines one adjustment value for forecast steps with the same judgments across covariates.
Relevant experience with matching covariate-wise judgments is retrieved based on similarity in the target context and base forecast.
Let $\mathcal{E}^{\mathrm{adj}}_w\subseteq\mathcal{M}_{<w}$ denote the retrieved relevant experience.
Its numerical adjustments provide references for determining the current adjustment.
The LLM then determines
\begin{equation}
a
=
g_\theta^{\mathrm{adj}}
\left(
y_{1:L},
X_w,
\hat{y}^{\mathrm{base}},
J,
\mathcal{E}^{\mathrm{adj}}_w
\right).
\end{equation}
The resulting adjustment is applied to the base forecast to obtain the forecast for the current window.

\subsection{Residual-Guided Experience Construction}

Once the target observation becomes available, \methodology{} revisits the forecast-time decision to construct validated experience for subsequent forecasts.
The error of the base forecast provides feedback on the numerical discrepancy from the target observation, but does not determine how the covariate effects should have been judged.
We represent this feedback by the observed residual.
At forecast step $h$, the observed residual is
\begin{equation}
r_h=
y_{L+h}-\hat{y}^{\mathrm{base}}_h,
\qquad
1\leq h\leq H.
\end{equation}
Let $r=(r_1,\ldots,r_H)\in\mathbb{R}^{H}$ denote the observed residual over the forecast horizon.

\paragraph{Alternative Judgment Generation.}

For reconstruction, $g_\theta^{\mathrm{alt}}$ uses the current context, base forecast, and observed residual $r$ to generate $N$ alternative covariate-wise judgments $\tilde{J}^{(1)},\ldots,\tilde{J}^{(N)} \in \mathcal{S}^{C\times H}$.
We exclude $J$ and $\mathcal{M}_{<w}$ from the input so that alternatives are generated without using the original forecast-time judgment or stored experience.
The original forecast-time judgment is then included with the generated alternatives for subsequent evaluation,
\begin{equation}
\mathcal{J}
=
\{J\}
\cup
\left\{
\tilde{J}^{(1)},\ldots,\tilde{J}^{(N)}
\right\}.
\end{equation}

\paragraph{Evaluation and Selection.}

Each judgment $J'\in\mathcal{J}$ is evaluated through a numerical adjustment produced by $g_\theta^{\mathrm{adj}}$ without stored experience.
The original judgment is re-evaluated under these conditions rather than using its forecast-time adjustment.
The resulting adjustment for $J'$ is
\begin{equation}
a(J')
=
g_\theta^{\mathrm{adj}}
\left(
y_{1:L},
X_w,
\hat{y}^{\mathrm{base}},
J'
\right).
\end{equation}
We compare the resulting adjustments with the observed residual and select the judgment whose adjustment best matches it.
Let $\mathcal{L}(r,a(J'))$ denote an error measure between the observed residual $r$ and the adjustment produced from $J'$.
The selected judgment and adjustment are
\begin{equation}
J^\star
=
\underset{J'\in\mathcal{J}}{\arg\min}
\; \mathcal{L}\left(r,a(J')\right),
\qquad
a^\star
=
a(J^\star).
\end{equation}

\paragraph{Validation and Memory Update.}

The selected adjustment may still fail to improve on the base forecast.
\methodology{} therefore validates the selected adjustment by comparing its error with the error of no adjustment, $\mathcal{L}(r,a^\star)<\mathcal{L}(r,0)$.
When this condition is satisfied, the selected judgment and adjustment are retained as validated experience together with their associated rationales, denoted by $R^{J,\star}$ and $R^{a,\star}$.
The validated experience for window $w$ is
\begin{equation}
e_w
=
\left(
y_{1:L},
X_w,
\hat{y}^{\mathrm{base}},
J^\star,
R^{J,\star},
a^\star,
R^{a,\star},
r
\right),
\qquad
\mathcal{M}_{<w+1}
\leftarrow
\mathcal{M}_{<w}\cup\{e_w\}.
\end{equation}
If the validation condition is not satisfied, the memory remains unchanged.
This validation does not establish that the selected judgment faithfully reflects the effects of the covariates.
Prompts for each LLM role are provided in Appendix~\ref{app:prompts}.

%% file: latex/400_experiments.tex
\subsection{Experimental Setup}

\paragraph{Datasets.}

We evaluate \methodology{} on diverse real-world datasets with covariates.
We consider five short-term electricity price forecasting datasets from the electricity price forecasting (EPF) benchmark~\citep{lago2021epf}.
This benchmark includes NP, PJM, BE, FR, and DE, which provide hourly electricity prices from different regional power markets together with two market-specific forecast covariates.
For long-term forecasting, we also consider ENTSO-e Load and Rossmann tasks from fev-bench~\citep{shchur2025fevbench}, which provide hourly electricity load with weather covariates and daily retail sales with calendar and promotion covariates, respectively.
Following MemCast~\citep{tao2026memcast}, we match its dataset sizes and reserve the final 20\% of each dataset for testing.
We use $L=7H$ throughout, with $H=24$, $168$, and $48$ for short-term forecasting, ENTSO-e, and Rossmann, respectively.
Detailed dataset descriptions and statistics are provided in Appendix~\ref{app:datasets}.

\paragraph{Baselines.}

We compare \methodology{} with representative statistical, training-based, and LLM-based forecasting methods.
Statistical baselines include ARIMA~\citep{hyndman2008arima} and Prophet~\citep{taylor2018prophet}, representing classical statistical forecasting approaches.
Training-based baselines include DLinear~\citep{zeng2023dlinear}, PatchTST~\citep{nie2023patchtst}, iTransformer~\citep{liu2024itransformer}, TimeXer~\citep{wang2024timexer}, and ConvTimeNet~\citep{cheng2025convtimenet}, spanning linear, Transformer-based, and convolutional forecasting architectures, together with Time-LLM~\citep{jin2024timellm}, which trains lightweight adaptation layers around a frozen LLM.
Inference-only LLM-based baselines include LSTPrompt~\citep{liu2024lstprompt}, LLM-Time~\citep{gruver2023llmtime}, and TimeReasoner~\citep{cheng2026timereasoner}, representing approaches that leverage LLMs for time series forecasting.
We further compare with the memory-based forecasting method MemCast~\citep{tao2026memcast}.
Detailed baseline descriptions are provided in Appendix~\ref{app:baselines}.

\paragraph{Implementation Details.}

We use Chronos-2~\citep{ansari2025chronos2} with its default inference configuration and 0.5 quantile as the frozen TSFM, and GPT-5 mini~\citep{openai2026gpt5} with \texttt{medium} reasoning effort as the LLM backbone.
The same GPT-5 mini configuration is used for LLM-based baselines with replaceable LLM backbones.
We generate $N=4$ alternative judgments and retrieve the top-$5$ experiences for both judgment formation and adjustment determination.
We evaluate forecasting accuracy using mean squared error (MSE) and mean absolute error (MAE), and use MSE as the error measure $\mathcal{L}$ for experience construction.
For each test window, MSE and MAE are computed over the $H$-step forecast horizon and then averaged across windows.
Lower values indicate better forecasting accuracy.
For the main experiments, $\mathcal{M}$ is accumulated chronologically over the training set and remains fixed over the test set.

\input{tables/tab_main_results}

\subsection{Main Results}

Table~\ref{tab:main_results} reports the overall forecasting performance across datasets.
\methodology{} achieves the lowest MSE and MAE on all seven datasets, outperforming the strongest baseline in the main evaluation for each dataset and metric by 16.7\% and 6.5\% on average, respectively.
The performance gap is particularly pronounced on the long-term datasets, where inference-only LLM forecasters are substantially less competitive.
Despite also relying on an LLM without task-specific training, \methodology{} achieves the lowest MSE and MAE on both ENTSO-e and Rossmann.

\paragraph{Dependence on Backbone Choice.}

\input{tables/tab_robust_backbone}

We examine whether the forecasting gains of \methodology{} persist across different TSFM and LLM backbones.
We evaluate TimesFM 2.5~\citep{das2024timesfm} and Toto 2.0-313M~\citep{khwaja2026toto2} as TSFM backbones with Qwen3.5-27B~\citep{qwen2026qwen35} and Gemini 3.5 Flash-Lite~\citep{google2026gemini35flash} as LLM backbones, reconstructing the experience memory independently for each combination.
As shown in Table~\ref{tab:robust_backbone}, all four backbone combinations outperform the second-best method in Table~\ref{tab:main_results} in MSE on both NP and PJM, while achieving comparable or better MAE.

\input{tables/tab_ablation_study}

\begin{figure}[t]
\noindent
\input{tables/tab_covariate_limit}
\hfill
\input{tables/tab_covariate_misalign}
\vspace{-8pt}
\end{figure}

\subsection{Ablation Study}

Table~\ref{tab:ablation_study} examines explicit covariate-wise judgment and experience construction in \methodology{}.

\paragraph{Explicit Judgment.}

Across the five short-term datasets, explicitly forming covariate-wise judgments before numerical adjustment reduces MSE over direct adjustment by 3.4\% without experience and 4.6\% with raw experience.
This improvement is less consistent on the long-term tasks, where forecast steps with the same covariate-wise judgments can require different numerical adjustments.
Beyond its forecast-time role, explicit judgment preserves the covariate-wise assessment for post-observation reconstruction.

\paragraph{Experience Construction.}
Relative to no experience, raw experience improves forecasting performance on some datasets but degrades it on others.
Retaining raw decisions only when their adjusted forecast improves on the base forecast performs better than raw experience on most datasets, but still does not consistently outperform no experience.
In contrast, validated experience outperforms no experience and both raw experience settings across datasets, reducing MSE and MAE by 12.1\% and 9.7\% on average relative to no experience.
This contrast shows that the gains from validated experience cannot be explained by validating the original decision alone, and instead depend on reconstructing the forecast-time decision after observation.

\subsection{Covariate Utilization Analysis}

Recent TSFMs such as Chronos-2 can effectively use covariate information across diverse real-world forecasting settings.
We refer to providing the available covariates directly to the TSFM together with the target context as direct covariate conditioning.
Its benefit, however, can be limited in some settings and sensitive to unreliable covariate information.
We examine whether \methodology{} can effectively use covariate information in these two settings.

\paragraph{Limited Direct Conditioning.}
\label{sec:covariate_limit}

We first examine whether \methodology{} can effectively use covariates when direct covariate conditioning provides limited forecasting benefit.
We consider three additional fev-bench tasks, UK COVID, Rohlik Orders, and M5, where direct conditioning provides no improvement over the target-only forecast.
Detailed dataset descriptions and experimental setups are provided in Appendix~\ref{app:covariate_limit}.
Table~\ref{tab:covariate_limit} shows that \methodology{} consistently improves the target-only forecast on all three tasks using the same covariates, whereas direct conditioning does not.
These results show that limited benefit from direct conditioning does not necessarily indicate limited forecasting utility of the covariates themselves.

\paragraph{Temporal Covariate Misalignment.}
\label{sec:covariate_misalign}

We next examine the sensitivity of \methodology{} to temporal misalignment in one covariate, with results shown in Table~\ref{tab:covariate_misalign}.
We apply 6 and 12-hour offsets to grid load on NP and PV wind on DE while leaving the other covariate unchanged, and use the same offset during experience construction and testing.
Although direct conditioning outperforms \methodology{} with aligned covariates, its forecasting error increases much more sharply under temporal misalignment.
Consequently, the advantage of direct conditioning is reversed at both offsets on NP and at 12 hours on DE.
The 0 judgment rate increases mainly for the shifted covariate while remaining relatively stable for the unchanged covariate.
Together, these results suggest that explicit covariate-wise assessment can limit the impact of unreliable covariates without discarding other covariate information.
Full MSE, MAE, and judgment statistics are provided in Appendix~\ref{app:covariate_misalign}.

\subsection{Experience Analysis}

We analyze how accumulating validated experience affects forecasting performance and retrieval, and whether its forecasting benefit depends on selecting relevant experience.
Detailed dataset-level results for experience accumulation, retrieval behavior, and relevant experience selection are provided in Appendices~\ref{app:experience_accumulation} and~\ref{app:experience_selection}.

\begin{figure}[t]
\centering
\includegraphics[width=\linewidth]{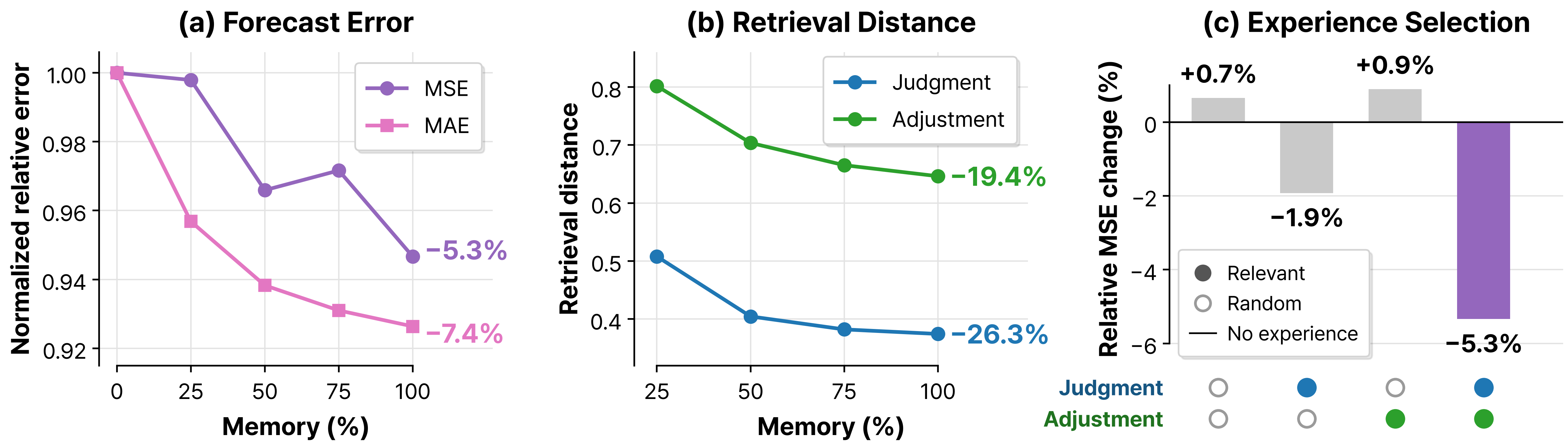}
\vspace{-10pt}
\caption{
Experience accumulation and retrieval analysis.
(a) Normalized forecasting error as validated experience accumulates, relative to no experience.
(b) Retrieval distance for judgment and adjustment retrieval across memory sizes.
(c) Relative MSE change under relevant and random retrieval at the judgment and adjustment stages using the full memory.
}
\label{fig:experience_analysis}
\vspace{-10pt}
\end{figure}

\paragraph{Experience Accumulation.}
\label{sec:experience_accumulation}

We first examine how forecasting performance changes as validated experience accumulates.
On the five short-term datasets, we build memory from the first 25\%, 50\%, 75\%, or 100\% of the training windows in chronological order, and evaluate each setting on the same test windows, including no experience.
Figure~\ref{fig:experience_analysis}(a) shows that forecasting performance improves overall as more validated experience accumulates, with full memory achieving the lowest MSE on all five datasets and reducing MSE and MAE by 5.3\% and 7.4\% on average relative to no experience.

\paragraph{Retrieval Behavior.}

Using the same memories evaluated in the preceding analysis, we next examine whether retrieved experience becomes more closely matched as validated experience accumulates.
For judgment and adjustment retrieval, we measure retrieval distance using the similarity criterion at each stage and average it over the top-$5$ retrieved experiences, test windows, and datasets.
Figure~\ref{fig:experience_analysis}(b) shows that as memory grows from 25\% to 100\%, average retrieval distance decreases from 0.508 to 0.374 for judgment retrieval and from 0.802 to 0.646 for adjustment retrieval.

\paragraph{Selecting Relevant Experience.}
\label{sec:experience_selection}

We finally examine whether the benefit of accumulated experience depends on selecting relevant experience at the judgment and adjustment stages.
We keep the full validated memory fixed and compare similarity-based and random retrieval independently at the two stages.
Figure~\ref{fig:experience_analysis}(c) shows that relevant judgment retrieval reduces MSE by 1.9\% even with random adjustment retrieval, whereas relevant adjustment retrieval alone does not improve over no experience.
Using relevant retrieval at both stages yields the largest average reductions in both MSE and MAE, with a 5.3\% MSE reduction and the lowest MSE among the four settings on all five datasets.
These results indicate that judgment retrieval plays the primary role by informing how covariate effects are assessed, while adjustment retrieval provides additional gains when determining the numerical adjustment from these assessments.

\begin{figure}[t]
\centering
\includegraphics[width=\linewidth]{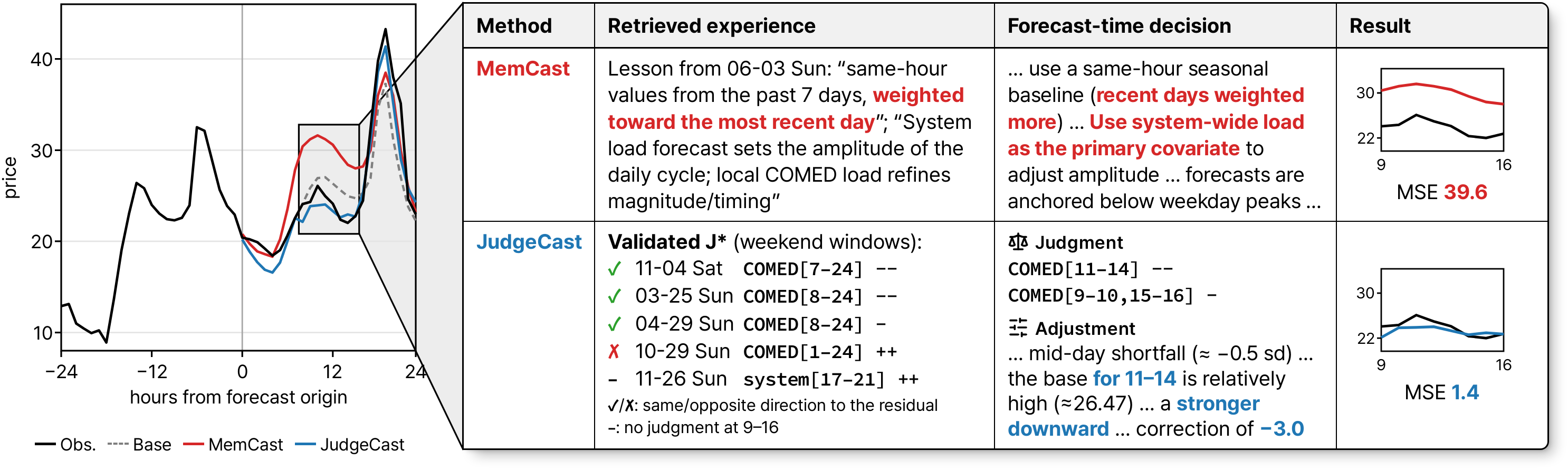}
\vspace{-5pt}
\caption{
PJM case study of \methodology{} and MemCast.
For the same window, retrieved experience informs each method's forecast-time decision and resulting forecast.
Only judgment retrieval is shown for \methodology{}.
Steps 9 to 16 highlight the interval where their forecasts differ most.
}
\label{fig:case_study}
\vspace{-10pt}
\end{figure}

\subsection{Case Study}

We use a PJM case study to examine how accumulated experience informs forecast-time decisions.
PJM targets the hourly day-ahead electricity price in the Commonwealth Edison zone, with system-wide and COMED day-ahead load forecasts as covariates.
Figure~\ref{fig:case_study} compares \methodology{} with MemCast, a memory-based forecaster that also accumulates experience, showing each method's retrieved experience, forecast-time decision, and forecast for the same window.

\paragraph{Forecast Behavior.}

We focus on steps 9 to 16, where the forecasts of \methodology{} and MemCast differ most.
Over this interval, \methodology{} applies a negative adjustment to the base forecast, reducing its overestimation, whereas MemCast substantially overestimates the observation.
The MSE over these steps is 1.4 for \methodology{} and 39.6 for MemCast.

\paragraph{Experience Use and Decision Formation.}

To understand this difference, we examine how each method uses retrieved experience at forecast time.
MemCast retrieves general forecasting guidance that places greater weight on recent same-hour values and uses system load as the primary covariate.
Following this guidance, its forecast-time decision prioritizes recent observations and system load, while the resulting forecast overestimates the observation over the highlighted interval.
In contrast, \methodology{} retrieves validated covariate-wise judgments from preceding weekend windows.
Most retrieved COMED judgments indicate a downward effect over similar daytime intervals, while one indicates the opposite direction.
Using this experience, \methodology{} assigns negative COMED judgments over steps 9 to 16 and determines downward numerical adjustments to the base forecast.
By retaining explicit forecast-time decisions as experience, \methodology{} allows the current judgment to be traced to the preceding decisions that informed it.

%% file: tables/tab_main_results.tex
\begin{table}[t]
\centering
\caption{
Overall forecasting performance across diverse real-world datasets.
Chronos-2 is the target-only frozen TSFM used as the base forecaster in \methodology{}.
Lower values indicate better performance.
The best and second-best results are shown in \textbf{bold} and \underline{underlined}, respectively.
\textsuperscript{a} and \textsuperscript{b} denote MSE and MAE reported in $\times 10^6$ and $\times 10^3$, respectively.
}
\label{tab:main_results}
\vspace{8pt}
\scriptsize
\setlength{\tabcolsep}{2.1pt}
\begin{tabular}{l|cc|cc|cc|cc|cc|cc|cc}
\toprule
\multirow{2}{*}{\textbf{Method}}
& \multicolumn{2}{c|}{NP} & \multicolumn{2}{c|}{PJM} & \multicolumn{2}{c|}{BE} & \multicolumn{2}{c|}{FR} & \multicolumn{2}{c|}{DE} & \multicolumn{2}{c|}{ENTSO-e} & \multicolumn{2}{c}{Rossmann} \\
& MSE & MAE & MSE & MAE & MSE & MAE & MSE & MAE & MSE & MAE & MSE\textsuperscript{a} & MAE\textsuperscript{b} & MSE\textsuperscript{a} & MAE\textsuperscript{b} \\
\midrule
ARIMA
& 45.681 & 4.108 & 53.988 & 5.217 & 1,319.458 & 16.622 & 1,391.465 & 12.591 & 334.941 & 11.841 & 46.943 & 3.161 & 8.760 & 2.065 \\
Prophet
& 50.383 & 5.086 & 70.194 & 6.379 & 950.769 & 17.790 & 1,003.981 & 14.898 & 464.086 & 15.197 & 7.544 & 1.986 & 10.793 & 2.228 \\
DLinear
& 34.315 & 3.817 & 41.687 & 4.640 & 758.796 & 12.426 & 792.836 & 9.619 & 236.175 & 10.537 & 3.573 & 1.301 & 10.210 & 2.248 \\
PatchTST
& 29.360 & 3.519 & 34.711 & 4.349 & 769.533 & 11.695 & 835.216 & 9.039 & 211.958 & 9.870 & 3.514 & 1.359 & 11.169 & 2.261 \\
iTransformer
& 30.941 & 3.537 & 35.467 & 4.240 & 767.894 & 12.195 & 943.188 & 11.195 & 247.900 & 10.857 & 3.548 & 1.380 & 11.220 & 2.354 \\
TimeXer
& 27.769 & 3.408 & \underline{29.458} & \underline{3.888} & 781.929 & 11.479 & 777.474 & 9.379 & 238.436 & 10.378 & 3.723 & 1.431 & 9.368 & 2.052 \\
ConvTimeNet
& 27.420 & 3.351 & 37.861 & 4.525 & 725.707 & 11.729 & 736.373 & 8.998 & 214.620 & 10.060 & 3.709 & 1.408 & 8.292 & 2.012 \\
Time-LLM
& 29.022 & 3.468 & 33.748 & 4.232 & 691.979 & 10.905 & 767.107 & 8.987 & 
\underline{211.507} & \underline{9.772} & 3.444 & 1.355 & 7.108 & 1.803 \\
LSTPrompt
& 41.120 & 4.084 & 46.662 & 4.906 & 743.230 & 13.336 & 986.232 & 10.597 & 333.517 & 12.202 & 18.996 & 2.895 & 10.683 & 2.107 \\
LLM-Time
& 73.374 & 5.873 & 111.900 & 7.277 & 1006.604 & 16.182 & 1045.227 & 13.062 & 677.279 & 14.771 & 45.379 & 5.499 & 15.714 & 2.790 \\
TimeReasoner
& 54.041 & 4.793 & 65.931 & 5.901 & 889.498 & 14.240 & 967.291 & 11.772 & 1,215.101 & 15.039 & 45.582 & 4.650 & 11.489 & 2.126 \\
MemCast
& 27.761 & 3.327 & 37.166 & 4.406 & 670.449 & 12.205 & 790.738 & 9.463 & 269.270 & 10.276 & 48.734 & 5.739 & 18.693 & 2.935 \\
\midrule
Chronos-2
& \underline{25.129} & \underline{2.990} & 36.641 & 4.333 & \underline{649.521} & \underline{10.272} & \underline{722.148} & \underline{7.559} & 224.197 & 9.786 & \underline{2.889} & \underline{1.093} & \underline{4.551} & \underline{1.311} \\
\rowcolor{gray!25} \textbf{\methodology}
& \textbf{19.657} & \textbf{2.875} & \textbf{25.946} & \textbf{3.736} & \textbf{610.917} & \textbf{9.971}
& \textbf{677.679} & \textbf{7.412} & \textbf{158.678} & \textbf{8.363} & \textbf{2.393} & \textbf{0.988} & \textbf{3.225} & \textbf{1.197} \\
\bottomrule
\end{tabular}
\vspace{-10pt}
\end{table}

%% file: tables/tab_robust_backbone.tex
\begin{wraptable}{r}{0.42\linewidth}
\vspace{-27pt}
\centering
\captionof{table}{
Backbone robustness on NP and PJM.
Qwen and Gemini abbreviate Qwen3.5-27B and Gemini 3.5 Flash-Lite.
TSFM-only rows report base forecasts.
}
\label{tab:robust_backbone}
\vspace{6pt}
\scriptsize
\setlength{\tabcolsep}{5.4pt}
\begin{tabular}{l|cc|cc}
\toprule
\multirow{2}{*}{\textbf{Backbone}}
& \multicolumn{2}{c|}{NP}
& \multicolumn{2}{c}{PJM} \\
& MSE & MAE & MSE & MAE \\
\midrule
TimesFM 2.5
& 23.655 & 3.079 & 33.135 & 4.106 \\
\; + Qwen
& 19.314 & 2.979 & 26.860 & 3.827 \\
\; + Gemini
& 17.637 & 2.829 & 28.503 & 3.884 \\
\midrule
Toto 2.0-313M
& 26.921 & 3.165 & 32.307 & 4.115 \\
\; + Qwen
& 21.831 & 3.092 & 24.905 & 3.673 \\
\; + Gemini
& 22.760 & 3.072 & 26.681 & 3.763 \\
\bottomrule
\end{tabular}
\end{wraptable}

%% file: tables/tab_ablation_study.tex
\begin{table}[t]
\centering
\caption{
Ablation study of explicit covariate-wise judgment and experience construction across datasets.
Raw experience retains the original forecast-time judgment or adjustment without post-observation reassessment.
Raw + Valid. retains the original judgment and adjustment only when the resulting adjusted forecast improves on the base forecast.
Validated experience applies residual-guided experience construction and corresponds to full \methodology{}.
Lower values indicate better performance.
\textsuperscript{a} and \textsuperscript{b} denote MSE and MAE reported in $\times 10^6$ and $\times 10^3$, respectively.
}
\label{tab:ablation_study}
\vspace{8pt}
\scriptsize
\setlength{\tabcolsep}{2pt}
\begin{tabular}{cc|cc|cc|cc|cc|cc|cc|cc}
\toprule
\multirow{2}{*}{\shortstack{\textbf{Explicit}\\\textbf{Judgment}}}
& \multirow{2}{*}{\shortstack{\textbf{Experience}\\\textbf{Construction}}}
& \multicolumn{2}{c|}{NP} & \multicolumn{2}{c|}{PJM}
& \multicolumn{2}{c|}{BE} & \multicolumn{2}{c|}{FR}
& \multicolumn{2}{c|}{DE} & \multicolumn{2}{c|}{ENTSO-e}
& \multicolumn{2}{c}{Rossmann} \\
& & MSE & MAE & MSE & MAE & MSE & MAE & MSE & MAE
& MSE & MAE & MSE\textsuperscript{a} & MAE\textsuperscript{b} & MSE\textsuperscript{a} & MAE\textsuperscript{b} \\
\midrule
$\times$ & None
& 22.551 & 3.206 & 30.493 & 4.012 & 624.713 & 10.003
& 689.195 & 7.461 & 178.935 & 8.699 & 3.128 & 1.209 & 4.469 & 1.249 \\
$\checkmark$ & None
& 21.514 & 3.096 & 30.290 & 4.112 & 619.435 & 10.842
& 680.784 & 8.250 & 161.724 & 8.564 & 3.359 & 1.201 & 4.556 & 1.386 \\
\midrule
$\times$ & Raw
& 22.422 & 2.908 & 34.429 & 4.219 & 641.466 & 10.310
& 718.869 & 7.564 & 210.147 & 9.421 & 2.773 & 1.068 & 4.363 & 1.247 \\
$\checkmark$ & Raw
& 22.767 & 2.990 & 29.975 & 3.958 & 623.891 & 10.158
& 712.784 & 7.710 & 192.831 & 9.012 & 3.162 & 1.195 & 7.194 & 1.649 \\
$\checkmark$ & Raw + Valid. & 21.961 & 2.884 & 29.855 & 3.952 & 615.480 & 10.028 & 701.613 & 7.812 & 176.561 & 8.720 & 3.216 & 1.180 & 7.255 & 1.632 \\
\midrule
\rowcolor{gray!25}
\textbf{$\checkmark$} & \textbf{Validated}
& \textbf{19.657} & \textbf{2.875} & \textbf{25.946} & \textbf{3.736} & \textbf{610.917} & \textbf{9.971}
& \textbf{677.679} & \textbf{7.412} & \textbf{158.678} & \textbf{8.363} & \textbf{2.393} & \textbf{0.988} & \textbf{3.225} & \textbf{1.197} \\
\bottomrule
\end{tabular}
\vspace{-10pt}
\end{table}

%% file: tables/tab_covariate_limit.tex
\begin{minipage}[t]{0.5\linewidth}
\centering
\captionof{table}{
Forecasting performance on fev-bench tasks where direct covariate conditioning with Chronos-2 does not improve the target-only forecast.
\textsuperscript{a} and \textsuperscript{b} denote MSE and MAE reported in $\times 10^6$ and $\times 10^3$, respectively.
}
\label{tab:covariate_limit}
\vspace{6pt}
\scriptsize
\setlength{\tabcolsep}{2.7pt}
\begin{tabular}{l|cc|cc|cc}
\toprule
\multirow{2}{*}{\textbf{Method}}
& \multicolumn{2}{c|}{UK COVID}
& \multicolumn{2}{c|}{Rohlik Orders}
& \multicolumn{2}{c}{M5} \\
& MSE\textsuperscript{a} & MAE\textsuperscript{b} & MSE\textsuperscript{a} & MAE & MSE & MAE \\
\midrule
Target-only
& 137.711 & 2.901 & 722.252 & 545.659 & 33.231 & 4.032 \\
Direct cov.
& 153.922 & 3.803 & 821.473 & 585.984 & 34.992 & 4.204 \\
\midrule
\rowcolor{gray!25}
\textbf{\methodology}
& \textbf{128.357} & \textbf{2.824} & \textbf{672.847} & \textbf{524.037} & \textbf{32.348} & \textbf{4.008} \\
\bottomrule
\end{tabular}
\end{minipage}

%% file: tables/tab_covariate_misalign.tex
\begin{minipage}[t]{0.46\linewidth}
\centering
\captionof{table}{
Forecasting performance on NP and DE with one temporally misaligned covariate.
Results are reported in MSE.
}
\label{tab:covariate_misalign}
\vspace{6pt}
\scriptsize
\setlength{\tabcolsep}{3.6pt}
\begin{tabular}{ll|ccc}
\toprule
\textbf{Dataset} & \textbf{Method} & 0h & 6h & 12h \\
\midrule
\multirow{3.7}{*}{\shortstack{NP\\(Grid load)}}
& Target-only & 25.129 & 25.129 & 25.129 \\
& Direct cov. & \textbf{15.899} & 23.695 & 25.043 \\
\cmidrule{2-5}
& \cellcolor{gray!25} \textbf{\methodology}
& \cellcolor{gray!25} 19.657 & \cellcolor{gray!25} \textbf{20.700} & \cellcolor{gray!25} \textbf{21.548} \\
\midrule
\multirow{3.7}{*}{\shortstack{DE\\(PV wind)}}
& Target-only & 224.197 & 224.197 & 224.197 \\
& Direct cov. & \textbf{73.164} & \textbf{131.504} & 195.862 \\
\cmidrule{2-5}
& \cellcolor{gray!25} \textbf{\methodology}
& \cellcolor{gray!25} 158.678 & \cellcolor{gray!25} 175.161 & \cellcolor{gray!25} \textbf{185.979} \\
\bottomrule
\end{tabular}
\end{minipage}

%% file: latex/500_conclusion.tex
We introduced \methodology{}, an experience-based framework for time series forecasting with covariates that makes covariate-wise judgments explicit.
At forecast time, \methodology{} uses accumulated experience to form these judgments and determine how the base forecast should be adjusted.
After observation, \methodology{} revisits the forecast-time decision to construct validated experience for subsequent forecasts.
Across diverse real-world datasets, \methodology{} consistently improves forecasting performance over strong baselines.
Our findings show that revisiting forecast-time decisions after observation can make forecast feedback reusable for subsequent decisions about covariate use.

%% file: latex/600_appendix.tex
\section{Experimental Details}

\subsection{Dataset Details}
\label{app:datasets}

Following MemCast~\citep{tao2026memcast}, we match its dataset sizes by taking the corresponding number of timestamps from the end of each dataset.
We reserve the final 20\% for testing.
For baselines that require training and validation, the preceding 80\% is divided chronologically into 70\% training and 10\% validation, preserving the same test set.
Table~\ref{tab:dataset_statistics} summarizes dataset statistics and temporal ranges used in our experiments.
Train and Test denote the number of timestamps per split.

\input{tables/tab_dataset_statistics}

\paragraph{Electricity Price Forecasting Datasets.}
NP, PJM, BE, FR, and DE are drawn from the electricity price forecasting (EPF) benchmark~\citep{lago2021epf}.
All five contain hourly day-ahead electricity prices together with two market-specific forecast covariates available before the target horizon.

\begin{itemize}
\item \textbf{NP}
represents the Nord Pool electricity market in the Nordic countries.
The target is the hourly day-ahead electricity price, with day-ahead load and wind generation forecasts as covariates.
\item \textbf{PJM}
represents the Pennsylvania, New Jersey, and Maryland electricity market in the United States.
The target is the hourly day-ahead price in the Commonwealth Edison zone, with system-wide and zonal day-ahead load forecasts as covariates.
\item \textbf{BE}
represents the Belgian electricity market.
The target is the hourly day-ahead Belgian electricity price, with French day-ahead load and generation forecasts as covariates.
\item \textbf{FR}
represents the French electricity market.
The target is the hourly day-ahead French electricity price, with day-ahead load and generation forecasts as covariates.
\item \textbf{DE}
represents the German electricity market.
The target is the hourly day-ahead German electricity price, with the Amprion zonal load forecast and aggregated wind and solar generation forecasts as covariates.
\end{itemize}

\paragraph{fev-bench Forecasting Tasks.}
fev-bench~\citep{shchur2025fevbench} is a benchmark for evaluating forecasters across diverse real-world settings and has been adopted in the evaluation of recent TSFMs.
It comprises 100 forecasting tasks across seven domains, including tasks with multivariate targets and covariates.
For long-term forecasting, we consider the 1-hour variant of ENTSO-e Load and the 1-day variant of Rossmann Store Sales.

\begin{itemize}
\item \textbf{ENTSO-e Load 1H}
contains hourly electricity load series from European countries.
Specifically, we use the Austrian series, with electricity load as the target and temperature, direct horizontal radiation, and diffuse horizontal radiation as weather covariates.
\item \textbf{Rossmann Store Sales 1D}
contains daily retail sales series with calendar and promotion covariates.
We evaluate a fixed random subset of 10 stores, shared across all methods.
\end{itemize}

\subsection{Baseline Details}
\label{app:baselines}

We provide detailed descriptions of the baselines used in our experiments below.

\paragraph{Statistical Methods.}

\begin{itemize}
\item \textbf{ARIMA}~\citep{hyndman2008arima}
combines autoregressive and moving-average components with differencing to model temporal dependencies in the target series.
\item \textbf{Prophet}~\citep{taylor2018prophet}
models a time series through additive trend, seasonality, and holiday components, with changepoints allowing the trend to vary over time.
\end{itemize}

\paragraph{Training-based Methods.}

\begin{itemize}
\item \textbf{DLinear}~\citep{zeng2023dlinear}
decomposes the input series into trend and remainder components and forecasts them with separate linear mappings before combining their outputs.
\item \textbf{PatchTST}~\citep{nie2023patchtst}
segments each univariate series into temporal patches and processes them with a channel-independent Transformer.
\item \textbf{iTransformer}~\citep{liu2024itransformer}
embeds each variable as a variate token and applies attention across variables to model multivariate dependencies.
\item \textbf{TimeXer}~\citep{wang2024timexer} represents the target at the patch level and covariates at the variate level, using global target tokens to integrate exogenous information into target forecasting.
\item \textbf{ConvTimeNet}~\citep{cheng2025convtimenet}
uses deformable patching and hierarchical convolutional blocks to capture local patterns and dependencies across multiple scales.
\item \textbf{Time-LLM}~\citep{jin2024timellm}
keeps the LLM backbone frozen while training reprogramming and projection layers to align time series patches with the LLM and produce forecasts.
\end{itemize}

\paragraph{LLM-based Methods.}

\begin{itemize}
\item \textbf{LSTPrompt}~\citep{liu2024lstprompt}
decomposes forecasting into short-term and long-term subtasks and prompts an off-the-shelf LLM with forecasting rules tailored to each subtask.
\item \textbf{LLM-Time}~\citep{gruver2023llmtime}
encodes numerical time series as strings of digits and formulates zero-shot forecasting as next-token prediction with a pretrained language model.
\item \textbf{TimeReasoner}~\citep{cheng2026timereasoner}
formulates forecasting as a conditional reasoning task and uses prompting strategies to elicit inference-time temporal reasoning from pretrained slow-thinking LLMs.
\item \textbf{MemCast}~\citep{tao2026memcast}
organizes accumulated forecasting experience into historical patterns, reasoning wisdom, and general laws, which guide later reasoning, trajectory selection, and reflective iteration.
It provides a direct comparison to \methodology{} as an LLM-based forecaster that also accumulates forecasting experience.
\end{itemize}

\subsection{Dataset Details for Limited Direct Conditioning Analysis}
\label{app:covariate_limit}

In addition to the fev-bench tasks used in the main evaluation, we consider three additional tasks for the limited direct conditioning analysis in Section~\ref{sec:covariate_limit}.
For this analysis, we set the context length to $L=7H$ for all tasks and retain the forecast horizon $H$ specified by fev-bench.
For the multivariate setting, we evaluate each target separately.

\begin{itemize}
\item \textbf{UK COVID Nation}
contains daily COVID-19 measurements for four UK nations, with new cases, deaths, and hospital admissions as targets and five covariates describing vaccination, hospitalization, and intensive-care occupancy.
We use $H=28$.
\item \textbf{Rohlik Orders}
contains daily order volumes from seven online-grocery warehouses, with 13 covariates describing calendar events, store operations, weather, and user activity.
We use $H=61$.

\item \textbf{M5 1D}
contains daily item-level Walmart sales with price, event, and SNAP indicators as covariates.
We do not use the static attributes as separate covariates.
We use $H=28$.
\end{itemize}

\section{Detailed Analysis Results}

\subsection{Temporal Covariate Misalignment Results}
\label{app:covariate_misalign}

Table~\ref{tab:app_covariate_misalign} reports the full results for the temporal covariate misalignment analysis in Section~\ref{sec:covariate_misalign}.
We report MSE and MAE at 0, 6, and 12 hour offsets, together with the covariate-wise rate of 0 judgments for the shifted and unchanged covariates in \methodology{}, computed as the proportion of forecast horizon timestamps assigned 0 across the test windows.

\input{tables/tab_covariate_misalign_appendix}
\input{tables/tab_experience_accumulation}

\subsection{Experience Accumulation and Retrieval Behavior Results}
\label{app:experience_accumulation}

Table~\ref{tab:experience_accumulation} reports the dataset-level results underlying the experience accumulation analysis in Section~\ref{sec:experience_accumulation}.
For experience accumulation, we report MSE and MAE as memory is constructed from 0\%, 25\%, 50\%, 75\%, and 100\% of the training windows, together with the number and proportion of constructed experiences retained as validated experience.
For retrieval behavior, we report the average retrieval distance for judgment and adjustment retrieval at each nonzero memory size.

\subsection{Selecting Relevant Experience Results}
\label{app:experience_selection}

Table~\ref{tab:experience_selection} reports the dataset-level results for the relevant experience selection analysis in Section~\ref{sec:experience_selection}.
Using the full validated memory, we compare similarity-based and random retrieval at the judgment and adjustment stages, including all four combinations.
For each setting, we report MSE and MAE together with the average retrieval distance at each stage.
The no experience setting is included as a reference, while relevant retrieval at both stages corresponds to the full \methodology{}.

\input{tables/tab_experience_selection}

\section{Qualitative Analysis}

\subsection{Case Visualization}

\begin{figure}[t]
\centering
\includegraphics[width=\linewidth]{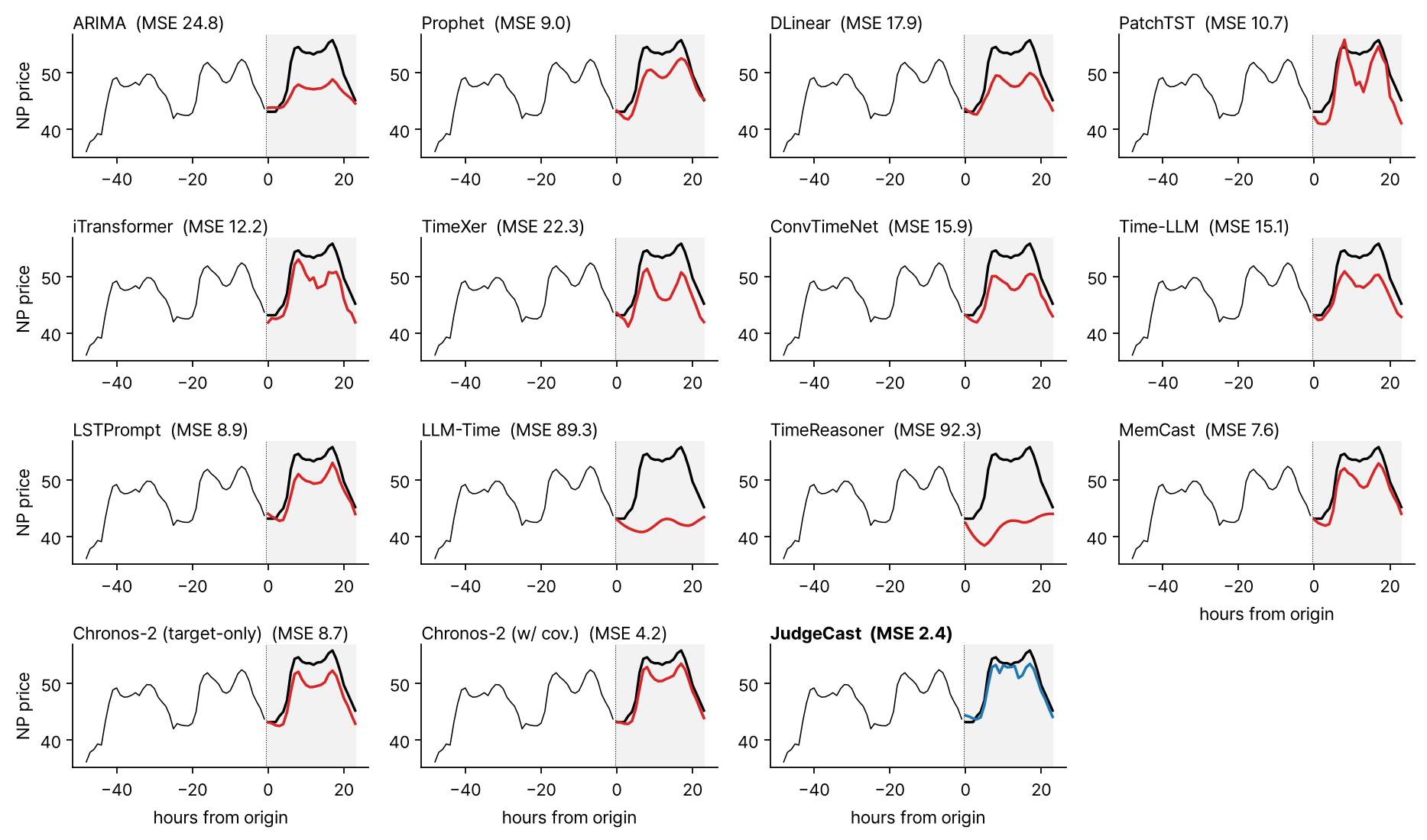}
\caption{
Case visualization on NP.
Most competing forecasts underestimate the sustained high price level, whereas \methodology{} adjusts upward and remains closer to the target observation.
}
\label{fig:case_vis_np}
\end{figure}

\begin{figure}[t]
\centering
\includegraphics[width=\linewidth]{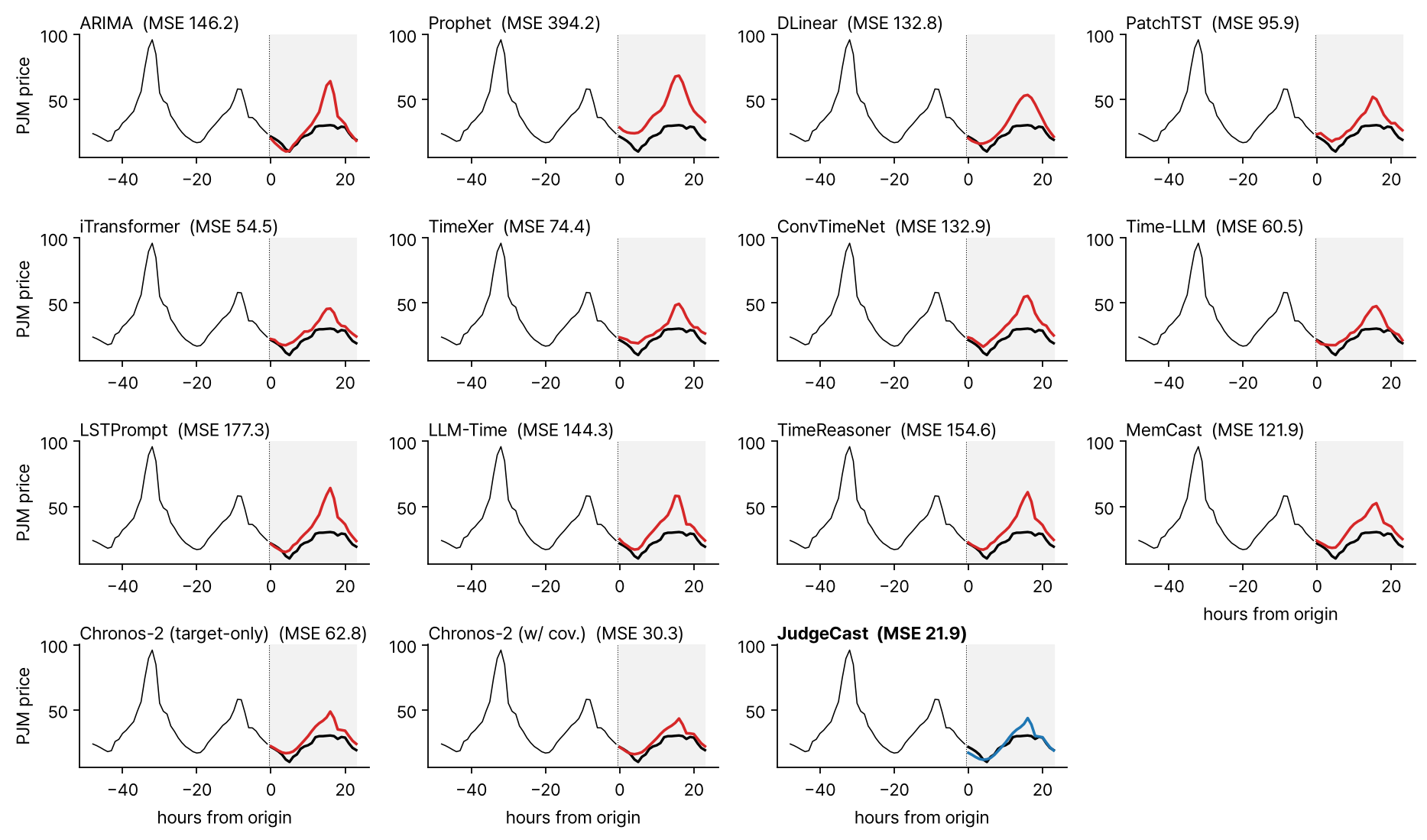}
\caption{
Case visualization on PJM.
Most competing forecasts overestimate the price peak, whereas \methodology{} reduces this overestimation and remains closer to the target observation.
}
\label{fig:case_vis_pjm}
\end{figure}

Figures~\ref{fig:case_vis_np} and~\ref{fig:case_vis_pjm} show two forecasting windows in which competing forecasts exhibit sustained errors in opposite directions.
In the NP case, most competing forecasts capture the initial price increase but underestimate the high price level sustained over the middle of the forecast horizon.
\methodology{} applies an upward adjustment to the base forecast over this interval, bringing it closer to the target observation.
In contrast, in the PJM case, most competing forecasts overestimate the subsequent price peak.
\methodology{} applies a downward adjustment to the base forecast over the forecast horizon, reducing this overestimation.
In both cases, \methodology{} achieves the lowest window-level MSE and MAE, outperforming direct covariate conditioning with Chronos-2, the second-best method.
Together, these cases illustrate that the gains of \methodology{} can arise from adjusting the base forecast in different directions depending on the forecasting context, rather than from consistently shifting it in a single direction.

\subsection{Failure Case Analysis}

\begin{figure}[t]
\centering
\includegraphics[width=\linewidth]{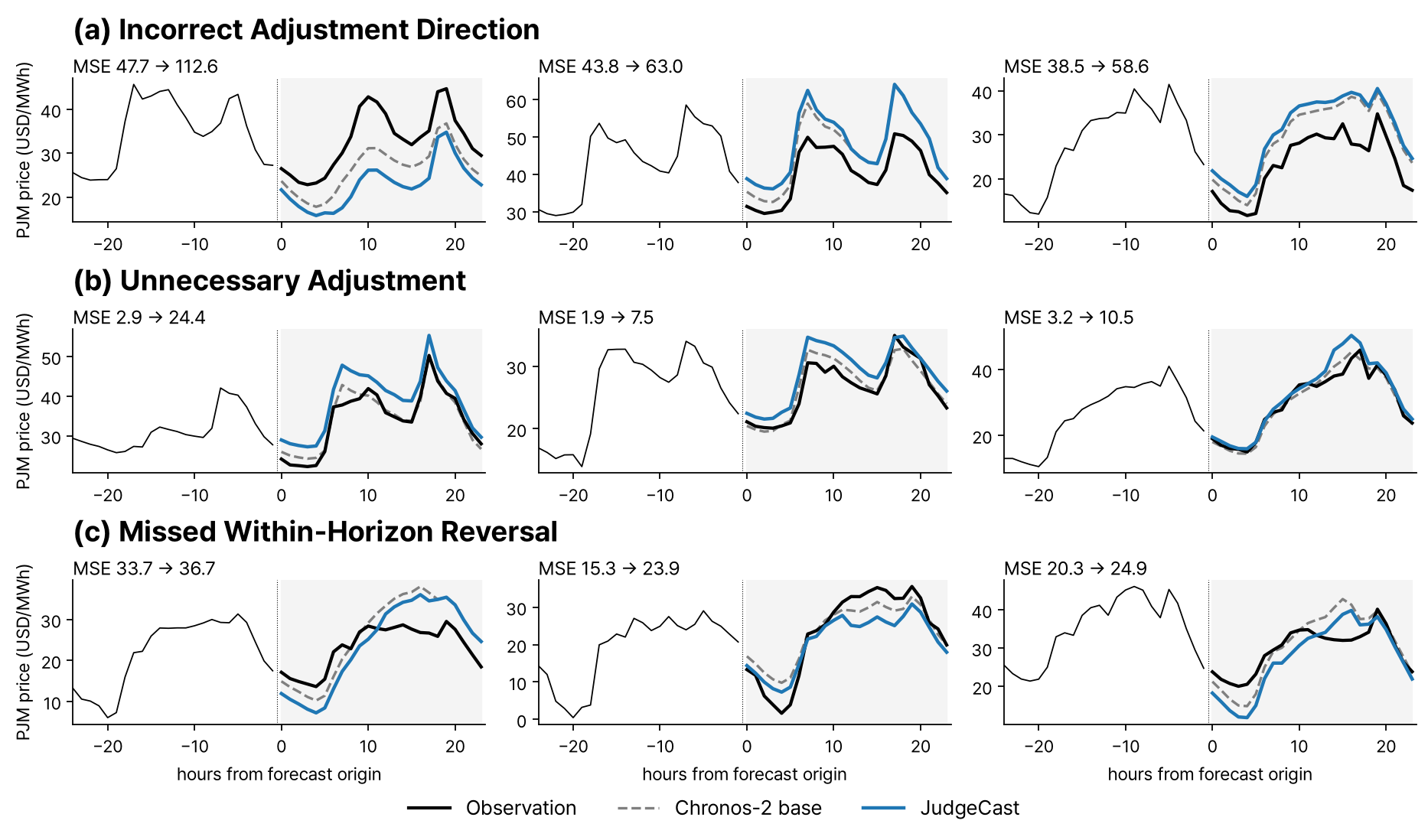}
\caption{
Representative failure cases of \methodology{} on PJM.
(a) Incorrect Adjustment Direction, where the numerical adjustment moves the base forecast in the direction opposite to the observed residual.
(b) Unnecessary Adjustment, where the base forecast is already close to the target observation but the numerical adjustment moves the final forecast away from it.
(c) Missed Within-Horizon Reversal, where the correction needed relative to the base forecast changes direction within the forecast horizon while the applied adjustment remains largely one-sided.
}
\label{fig:case_failure}
\end{figure}

We analyze failure cases of \methodology{} on PJM and identify three major failure modes in its forecast-time judgmental adjustment.
Figure~\ref{fig:case_failure} shows representative examples of each failure mode.

\paragraph{Incorrect Adjustment Direction.}

The numerical adjustment moves the base forecast in the direction opposite to the observed residual, increasing the forecast error.
Similar preceding windows generally support the selected adjustment direction, but some current spans exhibit the opposite residual direction. 
This discrepancy is more likely when the covariate departures are modest, with similar preceding windows also showing less consistent residual directions.
When \methodology{} nevertheless forms directional covariate-wise judgments and applies a substantial numerical adjustment in such spans, the resulting forecast can move considerably farther from the target observation.

\paragraph{Unnecessary Adjustment.}

The base forecast is already close to the target observation, but the numerical adjustment moves the final forecast away from it.
Similar covariate patterns in preceding windows generally support the applied adjustment, but the current window exhibits an unusually small observed residual.
This unusually small residual is not evident from the information available at forecast time.
As a result, an adjustment that is supported by the available context becomes unnecessary for the current window.

\paragraph{Missed Within-Horizon Reversal.}

The correction needed relative to the base forecast changes direction within the forecast horizon, while the applied adjustment remains largely one-sided.
The forecast-time covariate-wise judgments are also largely one-sided, following covariate departures that remain in the same direction over the horizon.
Similar preceding windows likewise do not consistently indicate a reversal in the observed residual.
The judgments thus fail to anticipate a change in the required correction that is not clearly indicated by the available forecast-time information.

\section{Prompts}
\label{app:prompts}

For reproducibility, we provide prompts for the three LLM roles in Figures~\ref{pmt:judgment} through~\ref{pmt:alternative}.

\input{figures/fig_prompts}

%% file: tables/tab_dataset_statistics.tex
\begin{table}[h]
\centering
\vspace{-10pt}
\caption{
Dataset statistics used in our experiments.
Rossmann uses a fixed random subset.
}
\label{tab:dataset_statistics}
\vspace{8pt}
\footnotesize
\setlength{\tabcolsep}{3.2pt}
\begin{tabular}{clccrrc}
\toprule
\textbf{Dataset}
& \textbf{Target}
& \textbf{\# Cov.}
& \textbf{Frequency}
& \textbf{Train}
& \textbf{Test}
& \textbf{Date Range} \\
\midrule
NP
& Nord Pool electricity price & 2 & 1 hour & 11,616 & 2,880 & 2017-04-30 -- 2018-12-24 \\
PJM
& COMED zonal electricity price & 2 & 1 hour & 11,616 & 2,880 & 2017-04-30 -- 2018-12-24 \\
BE
& Belgian electricity price & 2 & 1 hour & 11,616 & 2,880 & 2015-05-08 -- 2016-12-31 \\
FR
& French electricity price & 2 & 1 hour & 11,616 & 2,880 & 2015-05-08 -- 2016-12-31 \\
DE
& German electricity price & 2 & 1 hour & 11,616 & 2,880 & 2016-05-07 -- 2017-12-31 \\
ENTSO-e
& Austrian electricity load & 3 & 1 hour & 34,944 & 8,736 & 2015-01-01 -- 2019-12-26 \\
Rossmann
& Daily retail sales & 6 & 1 day & 7,530 & 1,890 & 2013-01-01 -- 2015-07-31 \\
\bottomrule
\end{tabular}
\vspace{-5pt}
\end{table}

%% file: tables/tab_covariate_misalign_appendix.tex
\begin{table}[h]
\centering
\caption{
Full results for temporal covariate misalignment.
Offsets are applied to grid load on NP and PV wind on DE while the other covariate remains unchanged.
Forecasting rows report MSE and MAE, and 0 Judgment rows report the covariate-wise rate of 0 judgments.
}
\label{tab:app_covariate_misalign}
\vspace{7pt}
\footnotesize
\begin{tabular}{ll|l|cc|cc|cc}
\toprule
\multirow{2}{*}{\textbf{Dataset}} & \multirow{2}{*}{\textbf{Result}} & \multirow{2}{*}{\textbf{\shortstack{Method /\\Covariate}}}
& \multicolumn{2}{c|}{\textbf{0h}}
& \multicolumn{2}{c|}{\textbf{6h}}
& \multicolumn{2}{c}{\textbf{12h}} \\
& &
& MSE & MAE & MSE & MAE & MSE & MAE \\
\midrule

\multirow{5}{*}{NP}
& \multirow{3}{*}{Forecasting}
& Target-only
& 25.129 & 2.990
& 25.129 & 2.990
& 25.129 & 2.990 \\

& & Direct cov.
& 15.899 & 2.336
& 23.695 & 2.848
& 25.043 & 2.998 \\

& & \methodology
& 19.657 & 2.875
& 20.700 & 2.991
& 21.548 & 3.069 \\

\cmidrule{2-9}

& \multirow{2}{*}{0 Judgment}
& Grid load
& \multicolumn{2}{c|}{23.6\%}
& \multicolumn{2}{c|}{26.7\%}
& \multicolumn{2}{c}{30.0\%} \\

& & Wind
& \multicolumn{2}{c|}{12.5\%}
& \multicolumn{2}{c|}{13.2\%}
& \multicolumn{2}{c}{11.5\%} \\

\midrule

\multirow{5}{*}{DE}
& \multirow{3}{*}{Forecasting}
& Target-only
& 224.197 & 9.786
& 224.197 & 9.786
& 224.197 & 9.786 \\

& & Direct cov.
& 73.164 & 5.428
& 131.504 & 7.437
& 195.862 & 9.044 \\

& & \methodology
& 158.678 & 8.363
& 175.161 & 8.907
& 185.979 & 9.399 \\

\cmidrule{2-9}

& \multirow{2}{*}{0 Judgment}
& PV wind
& \multicolumn{2}{c|}{17.2\%}
& \multicolumn{2}{c|}{29.8\%}
& \multicolumn{2}{c}{25.7\%} \\

& & Amprion load
& \multicolumn{2}{c|}{19.8\%}
& \multicolumn{2}{c|}{22.2\%}
& \multicolumn{2}{c}{21.7\%} \\

\bottomrule
\end{tabular}
\end{table}

%% file: tables/tab_experience_accumulation.tex
\begin{table}[h]
\centering
\caption{
Dataset-level results for experience accumulation and retrieval behavior.
Memory indicates the fraction of training windows used for experience construction.
The 0\% and 100\% settings denote no experience and full validated experience, respectively.
}
\label{tab:experience_accumulation}
\vspace{7pt}
\scriptsize
\setlength{\tabcolsep}{3.7pt}
\begin{tabular}{ll|cc|ccc|c|c}
\toprule
\multirow{2}{*}{\textbf{Dataset}}
& \multirow{2}{*}{\textbf{Memory}}
& \multicolumn{2}{c|}{\textbf{Forecasting}}
& \multicolumn{3}{c|}{\textbf{Experience Construction}}
& \textbf{Avg. Judgment}
& \textbf{Avg. Adjustment} \\
& & MSE & MAE
& \# Constructed & \# Validated & Valid. (\%)
& \textbf{Retrieval Distance} & \textbf{Retrieval Distance} \\
\midrule

\multirow{5}{*}{NP}
& 0\%   & 21.514 & 3.096 & 0   & 0   & --   & --    & --   \\
& 25\%  & 20.382 & 2.971 & 121 & 101 & 83.5 & 0.550 & 0.874 \\
& 50\%  & 19.857 & 2.871 & 238 & 202 & 84.9 & 0.484 & 0.774 \\
& 75\%  & 20.583 & 2.931 & 354 & 304 & 85.9 & 0.464 & 0.728 \\
& 100\% & 19.657 & 2.875 & 477 & 405 & 84.9 & 0.452 & 0.702 \\
\midrule

\multirow{5}{*}{PJM}
& 0\%   & 30.290 & 4.112 & 0   & 0   & --   & --    & --   \\
& 25\%  & 27.148 & 3.850 & 122 & 111 & 91.0 & 0.443 & 0.713 \\
& 50\%  & 26.823 & 3.810 & 239 & 222 & 92.9 & 0.327 & 0.622 \\
& 75\%  & 26.130 & 3.725 & 360 & 333 & 92.5 & 0.311 & 0.612 \\
& 100\% & 25.946 & 3.736 & 477 & 444 & 93.1 & 0.302 & 0.588 \\
\midrule

\multirow{5}{*}{BE}
& 0\%   & 619.435 & 10.842 & 0   & 0   & --   & --    & --   \\
& 25\%  & 620.595 & 10.469 & 121 & 109 & 90.1 & 0.495 & 0.805 \\
& 50\%  & 620.328 & 10.182 & 247 & 218 & 88.3 & 0.390 & 0.702 \\
& 75\%  & 624.832 & 10.247 & 363 & 328 & 90.4 & 0.368 & 0.663 \\
& 100\% & 610.917 & 9.971 & 477 & 437 & 91.6 & 0.360 & 0.633 \\
\midrule

\multirow{5}{*}{FR}
& 0\%   & 680.784 & 8.250 & 0   & 0   & --   & --    & --   \\
& 25\%  & 701.284 & 7.423 & 118 & 109 & 92.4 & 0.465 & 0.726 \\
& 50\%  & 696.773 & 7.477 & 236 & 218 & 92.4 & 0.368 & 0.632 \\
& 75\%  & 684.141 & 7.292 & 353 & 326 & 92.4 & 0.348 & 0.591 \\
& 100\% & 677.679 & 7.412 & 477 & 435 & 91.2 & 0.344 & 0.593 \\
\midrule

\multirow{5}{*}{DE}
& 0\%   & 161.724 & 8.564 & 0   & 0   & --   & --    & --   \\
& 25\%  & 180.149 & 8.760 & 119 & 112 & 94.1 & 0.587 & 0.890 \\
& 50\%  & 161.097 & 8.496 & 239 & 224 & 93.7 & 0.454 & 0.788 \\
& 75\%  & 165.796 & 8.336 & 359 & 337 & 93.9 & 0.420 & 0.732 \\
& 100\% & 158.678 & 8.363 & 477 & 449 & 94.1 & 0.414 & 0.715 \\
\bottomrule
\end{tabular}
\end{table}

%% file: tables/tab_experience_selection.tex
\begin{table}[h]
\centering
\caption{
Dataset-level results for selecting relevant experience.
Retrieval denotes the strategies used for judgment and adjustment retrieval, respectively.
Relevant uses the similarity-based retrieval criterion of \methodology{}, while Random replaces it with random selection.
All retrieval settings use the full validated memory, and Relevant / Relevant corresponds to \methodology{}.
}
\label{tab:experience_selection}
\vspace{7pt}
\small
\begin{tabular}{ll|cc|c|c}
\toprule
\multirow{2}{*}{\textbf{Dataset}}
& \multirow{2}{*}{\textbf{Retrieval}}
& \multicolumn{2}{c|}{\textbf{Forecasting}}
& \textbf{Judgment}
& \textbf{Adjustment} \\
& & MSE & MAE
& Retr. Dist.
& Retr. Dist. \\
\midrule

\multirow{5}{*}{NP}
& No Experience         & 21.514 & 3.096 & --    & --    \\
& Random / Random       & 20.974 & 3.003 & 1.154 & 1.035 \\
& Relevant / Random     & 20.576 & 3.001 & 0.452 & 1.037 \\
& Random / Relevant     & 21.469 & 2.996 & 1.154 & 0.692 \\
& Relevant / Relevant   & 19.657 & 2.875 & 0.452 & 0.702 \\
\midrule

\multirow{5}{*}{PJM}
& No Experience         & 30.290 & 4.112 & --    & --    \\
& Random / Random       & 29.702 & 4.008 & 0.952 & 0.903 \\
& Relevant / Random     & 26.500 & 3.820 & 0.302 & 0.905 \\
& Random / Relevant     & 28.498 & 3.927 & 0.952 & 0.584 \\
& Relevant / Relevant   & 25.946 & 3.736 & 0.302 & 0.588 \\
\midrule

\multirow{5}{*}{BE}
& No Experience         & 619.435 & 10.842 & --    & --    \\
& Random / Random       & 633.665 & 10.273 & 0.988 & 0.958 \\
& Relevant / Random     & 625.879 & 10.157 & 0.360 & 0.940 \\
& Random / Relevant     & 630.511 & 10.316 & 0.988 & 0.648 \\
& Relevant / Relevant   & 610.917 & 9.971 & 0.360 & 0.633 \\
\midrule

\multirow{5}{*}{FR}
& No Experience         & 680.784 & 8.250 & --    & --    \\
& Random / Random       & 681.604 & 7.375 & 0.986 & 0.903 \\
& Relevant / Random     & 690.065 & 7.469 & 0.344 & 0.891 \\
& Random / Relevant     & 699.525 & 7.329 & 0.986 & 0.588 \\
& Relevant / Relevant   & 677.679 & 7.412 & 0.344 & 0.593 \\
\midrule

\multirow{5}{*}{DE}
& No Experience         & 161.724 & 8.564 & --    & --    \\
& Random / Random       & 170.344 & 8.423 & 1.083 & 1.038 \\
& Relevant / Random     & 169.595 & 8.514 & 0.414 & 1.043 \\
& Random / Relevant     & 171.544 & 8.587 & 1.083 & 0.720 \\
& Relevant / Relevant   & 158.678 & 8.363 & 0.414 & 0.715 \\
\bottomrule
\end{tabular}
\end{table}

%% file: figures/fig_prompts.tex
\begin{figure}[t]
\centering
\begin{promptbox}{Covariate-wise Judgment Formation}
You are the covariate judge. A frozen forecaster has already produced a base forecast for {target_name} using only the target's own history. For each covariate, identify the horizon spans, if any, where it should push the target below or above the base forecast, and judge how strongly. Leave steps without a supported effect at 0.

# Inputs
Target: {target_name}
Forecast steps: 1-{horizon}
Judgment labels: {judgments}

Target history and current target state:
{target_context}

Target-only base forecast:
{y_base}

Covariates available at forecast time, with their descriptions and values:
{covariate_block}

Forecast-time situation derived from the information above, including the covariates' current/seasonal departures and the target/base position:
{perception_block}

Retrieved experiences from contextually similar past windows:
{memory_block}

# Output
Return exactly one JSON object. List only nonzero judged spans in "judgments". Every covariate must appear exactly once in "rationales".

{
  "judgments": [
    {"covariate": "name", "start": 1, "end": 1, "judgment": "+"}
  ],
  "rationales": [
    {"covariate": "name", "rationale": "why this covariate receives its judgment(s), or remains at 0, in the current window"}
  ]
}
\end{promptbox}
\caption{Prompt for forecast-time covariate-wise judgment formation.}
\label{pmt:judgment}
\end{figure}

\begin{figure}[t]
\centering
\begin{promptbox}{Numerical Adjustment Determination}
You are the adjuster. The covariate judgments for this window are fixed. Horizon steps with the same joint judgment pattern form one group. For each group, choose one numerical correction in {target_name}'s units; the same correction is applied to every step in that group.

# Inputs
Target: {target_name}
Groups to size:
{group_ids}

Current target/base context:
{current_profile}

Current target scale (standard deviation of the target over the lookback):
{current_target_scale}

Groups. Each group contains its fixed joint judgment with the judge's rationale for each member, the current-window situation on its steps, and retrieved past cases with the same joint judgment pattern. A past case may include its situation, its source target scale, its relative adjustment magnitude, the adjustment selected there when the scales are comparable, and the stored adjustment rationale. Relative adjustment magnitude is the absolute historical adjustment divided by the source target scale:
{groups}

# Output
Return exactly one JSON object with one entry for every group id.

{
  "adjustments": [
    {"id": "g0", "delta": 0.0, "rationale": "current-window grounds for this correction"}
  ]
}
\end{promptbox}
\caption{Prompt for forecast-time numerical adjustment determination.}
\label{pmt:adjustment}
\end{figure}

\begin{figure}[t]
\centering
\begin{promptbox}{Alternative Judgment Generation}
You are proposing alternative covariate judgments for a completed forecasting window. The target observation has arrived, so the base residual is known. The residual is aggregate feedback: it shows where and by how much the base forecast was wrong, but it does not identify which covariate caused the error.

# Inputs
Target: {target_name}
Forecast steps: 1-{horizon}
Judgment labels: "--" | "-" | "0" | "+" | "++"

Target history and target state that were available at forecast time:
{target_context}

Target-only base forecast:
{y_base}

Covariates that were available at forecast time, with their descriptions and values:
{covariate_block}

Forecast-time situation derived from those inputs, including the covariates' current/seasonal departures and the target/base position:
{perception_block}

Observed base residual, step by step. Positive means the observation was above the base forecast; negative means it was below:
{residual_block}

# Output
Return exactly one JSON object with exactly {n_candidates} distinct candidates, in ranked order.

{
  "candidates": [
    {
      "judgments": [
        {"covariate": "name", "start": 1, "end": 1, "judgment": "+"}
      ],
      "rationales": [
        {"covariate": "name", "rationale": "forecast-time evidence for this covariate's judgment(s), or for leaving it at 0"}
      ]
    }
  ]
}
\end{promptbox}
\caption{Prompt for post-observation alternative judgment generation.}
\label{pmt:alternative}
\end{figure}